%% file: csgs_edera-strappa-bromberg.tex
\documentclass[runningheads,a4paper,orbibl]{llncs}
\usepackage[lined,ruled,linesnumbered,boxed,noend]{algorithm2e}
\usepackage{amssymb}
\usepackage{amsopn}
\usepackage{graphicx}

\usepackage{url}
\urldef{\mailsa}\path|{aedera, ystrappa, fbromberg}@frm.utn.edu.ar|
\newcommand{\keywords}[1]{\par\addvspace\baselineskip
\noindent\keywordname\enspace\ignorespaces#1}

\newcommand{\ci}[3]{\ensuremath{I({#1},{#2} \mid #3)}}
\newcommand{\G}{\ensuremath{\mathcal{G}}}
\newcommand{\F}{\ensuremath{\mathcal{F}}}
\newcommand{\D}{\ensuremath{\mathcal{D}}}
\newcommand{\X}{\ensuremath{\mathcal{X}}}
\newcommand{\BigO}[1]{\ensuremath{\operatorname{O}\bigl(#1\bigr)}}
\DeclareMathOperator{\val}{Val}
\DeclareMathOperator{\MB}{MB}
\DeclareMathOperator*{\argmax}{\arg\!\max}
\newcommand{\mb}[1]{\ensuremath{\MB({#1})}}
\newcommand{\mbg}[2]{\ensuremath{\MB({#1~\colon~#2})}}

\newcommand{\currentTitle}{The Grow-Shrink strategy for learning Markov
  network structures constrained by context-specific independences}
\newcommand{\currentAuthors}{Alejandro Edera\and Yanela Strappa\and Facundo
  Bromberg}

\begin{document}

\mainmatter  

\title{\currentTitle}

\titlerunning{The Grow-Shrink strategy for learning Markov network structures}

 \author{Alejandro Edera\and Yanela Strappa\and Facundo Bromberg}
 \authorrunning{\currentAuthors}

\institute{Departamento de Sistemas de Informaci\'on, Universidad Tecnol\'ogica Nacional,\\
   Rodriguez 273, M5502 Mendoza, Argentina\\
   \mailsa}

\maketitle

\begin{abstract}

  Markov networks are models for compactly representing complex probability
  distributions. They are composed by a \emph{structure} and a set of
  numerical weights. The structure qualitatively describes independences in
  the distribution, which can be exploited to factorize the distribution into
  a set of compact functions. A key application for learning structures from
  data is to automatically discover knowledge. In practice, structure learning
  algorithms focused on ``\emph{knowledge discovery}'' present a limitation:
  they use a coarse-grained representation of the structure. As a result, this
  representation cannot describe \emph{context-specific independences}. Very
  recently, an algorithm called CSPC was designed to overcome this limitation,
  but it has a high computational complexity. This work tries to mitigate this
  downside presenting CSGS, an algorithm that uses the Grow-Shrink strategy
  for reducing unnecessary computations. On an empirical evaluation, the
  structures learned by CSGS achieve competitive accuracies and lower
  computational complexity with respect to those obtained by CSPC.

\end{abstract}

\section{Introduction}

Markov networks are parametric models for compactly representing complex
probability distributions of a wide variety of domains. These models are
composed by two elements: a \emph{structure} and a set of \emph{numerical
  weights}. The structure plays an important role, because it describes a set
of independences that holds in the domain, thus making assumptions about the
functional form or factorization of the
distribution~\cite{Hammersley_Clifford_1971}. 
For this reason, the structure is an important source of knowledge discovery
because it depicts intricate patterns of probabilistic (in)dependences between
the domain variables. 
%
Usually, the structure of a Markov network can be constructed by algorithms
using observations taken from an unknown distribution. Interestingly, the
constructed structure can be used by human experts for discovering unknown
knowledge
\cite{smith2006computational}. 
For this reason, the problem of structure learning from data has received an
increasing attention in machine learning
\cite{pearl88,lauritzen1996graphical,koller09}. However, Markov network
structure learning from data is still challenging. One of the most important
problems is that it requires weight learning that cannot be solved in
closed-form, requiring to perform a convex optimization with inference as a
subroutine. Unfortunately, inference in Markov networks is \#P-complete
\cite{koller09}.

As a result, structure learning algorithms seek the ``best'' approximation to
the solution structure, making assumptions about the form of the solution
space or the used objective function. The choice of these approximations
depends on the \emph{goal of learning} used for designing learning algorithms
\cite[Chapter~16]{koller09}. In generative learning, we can find two goals of
learning: \emph{density estimation}, where a structure is ``best'' when the
resulting Markov network is \emph{accurate} for answering inference queries;
and \emph{knowledge discovery}, where a structure is ``best'' when it is
\emph{accurate} for qualitatively describing the independences that hold in
the distribution. Depending on the goal of learning, we can categorize
structure learning algorithms in: \emph{density estimation algorithms}
\cite{van2012markov,JMLR:v15:lowd14a}; and \emph{knowledge discovery
  algorithms} \cite{bromberg2009efficient,schluter2014ibmap}. In this work, we
are focusing in the knowledge discovery goal.

In practice, knowledge discovery algorithms exploit the fact that the
structure can be viewed as a set of independences. Thus, for constructing a
structure, such algorithms successively make \emph{(in)dependence queries} to
data in order to restrict the number of possible structures, converging toward
the solution structure. To achieve a good performance in this procedure,
knowledge discovery algorithms use a sound and complete \emph{representation
  of the structure}: a single undirected graph. A graph can be viewed as an
inference engine which efficiently represents and manipulates (in)dependences
in polynomial time \cite{pearl88}.  Unfortunately, this graph representation
cannot capture a type of independences known as \emph{context-specific
  independences}
\cite{hojsgaard2000yggdrasil,hojsgaard2004statistical,koller09}. For these
cases, knowledge discovery algorithms cannot achieve good results in their
goal of learning, because a single graph cannot capture such independences,
obscuring the acquisition of knowledge. To overcome this limitation, a novel
knowledge discovery algorithm has recently been developed
\cite{edera2014a}. This algorithm, called CSPC, uses an alternative
representation of the structure called \emph{canonical models}, a particular
class of \emph{Context Specific Interaction models} (CSI models)
\cite{hojsgaard2004statistical}. Canonical models allow us to encode
context-specific independences by using a set of mutually independent
graphs. Using this representation, CSPC can learn more accurate structures
than several state-of-the-art algorithms. However, despite the benefits in
accuracy, CSPC presents an important downside: it has a high computational
complexity, because it must perform a large number of independence queries in
comparison to traditional algorithms.

Therefore, this paper focuses on reducing the number of independence queries
required for learning canonical models. This reduction was thought in order to
achieve competitive accuracies with respect to CSPC, but avoiding unnecessary
queries. To achieve this, we present the CSGS algorithm, a knowledge discovery
algorithm that learns canonical models by using the Grow-Shrink
strategy~\cite{margaritis00} in a similar way to the GSMN algorithm, a
Markov network structure learning
algorithm~\cite{bromberg2009efficient}. Basically, under the assumption of
bounded maximum degree, this strategy constructs a structure in polynomial
time by identifying local neighborhoods of each
variable~\cite{margaritis00}. On an empirical evaluation, the canonical models
learned by CSGS achieve competitive accuracies and lower time complexity with
respect to those obtained by CSPC. 

The remaining of this work is structured as follows:
Section~\ref{sec:background} reviews essential
concepts. Section~\ref{sec:csgs} presents our contribution: CSGS. Next,
Section~\ref{sec:empirical-evaluation} shows our empirical evaluation of CSGS
on synthetic datasets. Finally, Section~\ref{sec:concl-future-work} concludes
with directions for future work.

\section{Background}
\label{sec:background}

We introduce our general notation. Hereon, we use the symbol \(V\) to denote a
finite set of indexes. Lowercase subscripts denote particular indexes, for
instance \(a, b\in V\); in contrast, uppercase subscripts denote subsets of
indexes, for instance \(W\subseteq V\). Let \(X_V\) be a set of random
variables of a domain, where single variables are denoted by single indexes in
\(V\), for instance \(X_a, X_b\in X_V\) where \(a,b\in V\). We simply use
\(X\) instead of \(X_V\) when \(V\) is clear from the context. We focus on the
case where \(X\) takes discrete values \(x\in\val(V)\), that is, the values
for any \(X_a\in X\) are discrete: \(\val(a)=\{x_a^0, x_a^1, \ldots\}\). For
instance, for boolean-valued variables, that is \(|\val(a)|=2\), the symbols
\(x^0_a\) and \(x^1_a\) denote the assignments \(X_a=0\) and \(X_a=1\),
respectively. Moreover, we overload the symbol \(V\) to also denote the set of
nodes of a graph. Finally, we use \(\X\subseteq\val(V)\) for denoting an
arbitrary set of complete or \emph{canonical assignments}, that is, all the
variables take a fixed value. For instance, \(x_V^i \equiv x^i\in\val(V)\).

\subsection{Conditional and context-specific independences}
\label{sec:cont-spec-cond}

A set of independence assumptions is commonly called the \emph{structure} of a
distribution because independences determine the factorization, or functional
form, of a distribution. Two of the most known types of independences are
conditional and context-specific independences. The latter has received an
increased interest
\cite{hojsgaard2000yggdrasil,hojsgaard2004statistical,koller09,edera2013a,edera2014a},
because one conditional independence can be expressed as a set of
context-specific independences. Formally, context-specific independences are
defined as follows:

\begin{definition}
  \label{def:cond-cont-spec}
  Let \(A , B, U, W \subseteq V\) be disjoint subsets of indexes, and let
  $x_W$ be some assignment in \(\val(W)\).  Let \(p(X)\) be a probability
  distribution. We say that variables \(X_A\) and \(X_B\) are
  \emph{contextually independent} given \(X_U\) and the context $X_W=x_W$,
  denoted by $\ci{X_A}{X_B}{X_U,x_W}$, iff \(p(X)\) satisfies:
  \[ p(x_A \vert x_B, x_U, x_W) = p(x_A
    \vert x_U, x_W),\]
  for all assignments \(x_A\), \(x_B\), and \(x_U\); whenever \(p(x_B, x_U,
  x_W) > 0\).
\end{definition}

As a consequence, if \ci{X_A}{X_B}{X_U,x_W} holds in \(p(X)\), then it
logically follows that \ci{x_A}{x_B}{x_U, x_W} also holds in \(p(X)\) for
any assignment \(x_A\), \(x_B\), \(x_U\). Interestingly, if
\ci{X_A}{X_B}{X_U,x_W} holds for all \(x_W\in\val(W)\), then we say that the
variables are conditionally independent. Formally,

\begin{definition}
  \label{def:cond}
  Let \(A , B, U, W \subseteq V\) be disjoint subsets of indexes, and let
  \(p(X)\) be a probability distribution. We say that variables \(X_A\) and
  \(X_B\) are \emph{conditionally independent} given \(X_U\) and \(X_W\),
  denoted by \ci{X_a}{X_b}{X_U,X_W}, iff \(p(X)\) satisfies:
  \[
  p(x_A \vert x_B, x_U, x_W) = p(x_A \vert x_U, x_W),
  \]
    for all assignments \(x_A\), \(x_B\), \(x_U\), and \(x_W\); whenever
    \(p(x_B, x_U, x_W)
    > 0\).
\end{definition}

Thus, a conditional independence \ci{X_A}{X_B}{X_U, X_W} that holds in
\(p(X)\) can be seen as a conjunction of context-specific independences of the
form \(\bigwedge_{x_W}\ci{X_A}{X_B}{X_U, x_W}\) for all
\(x_W\in\val(W)\). Moreover, each context-specific independence
\ci{X_A}{X_B}{X_U, x_W}, that holds in \(p(X)\), can be seen as a conditional
independence \ci{X_A}{X_B}{X_U} that holds in the conditional distribution
\(p(X_{V\setminus W}|x_W)\)\cite{edera2013a}.



\subsection{Representation of structures}
\label{sec:repr-struct}

The independence relation \ci{\cdot}{\cdot}{\cdot} commonly assumes \emph{the
  Markov properties}~\cite[Section 3.1]{lauritzen1996graphical}; we also
assume that probability distributions are \emph{positive}\footnote{A
  distribution \(p(X)\) is positive if \(p(x) > 0\), for all
  \(x\in\val(V)\).}. Thus, an isomorphic mathematical object that conforms to
the previous properties is an undirected graph \cite{pearl88}. An undirected
graph \(G\) is a pair \((V, E)\), where \(E\subset V\times V\) is a set of
edges which encodes conditional independences by using the graph-theoretic
notion of \emph{reachability}. As a result, the independence assertion
\ci{X_A}{X_B}{X_U} can be associated with the graphical condition: ``every
path from A to B is intercepted by the nodes U''. Therefore, a graph \(G\)
encodes knowledge in a readily accessible way, that is, the graph is highly
interpretable. For instance, we can determine the adjacencies of a node \(a\in
V\), or its \emph{Markov blanket} \(\mbg{a}{G}\subseteq
V\setminus\{a\}\)\footnote{We simply use \(\mb{a}\) when the structure from
  which the Markov blanket is defined is clear from the context.}, from its
neighboring nodes in the graph \(G\). Unfortunately, the use of a single graph
as representation presents an issue when distributions hold context-specific
independences, because it only encodes conditional independences, leading to
excessively dense graphs \cite{edera2013a,edera2014a}.

In practice, for overcoming the previous limitation, an alternative
representation of the structure consists in a set \(\F=\{f^i_D\}\) of
features, where each feature is commonly represented as an indicator function
(Kronecker's delta), that is, a boolean-valued function
\(f_D:\val(D)\mapsto\{0,1\}\). Given an arbitrary assignment \(x\), a feature
\(f^i_D(x)\) returns 1, if \(x_D = x^i_D\); and \(0\) otherwise. A set of
features is a more flexible representation than a graph, because the former
can encode context-specific independences. For example, an independence of the
form \ci{X_a}{X_b}{x_W} is encoded in \F\ iff for any feature
\(f^i_D\in\F'=\{f^i_D~\in~\F~\colon~x_W = x^i_W~\wedge~W\subseteq D\}\), the
variables \(X_a\) and \(X_b\) do not appear simultaneously in the set \(D\),
that is, either \(a\notin D\) or \(b\notin D\). From a set \(\F\) of features,
we can induce a graph \(G\) by adding an edge between every pair of nodes
whose variables appear together in some feature \(f^i_D\in\F\)
\cite{edera2014a}. In a similar way, following our previous example, we can
induce a graph from \(\F'\subseteq\F\). This graph is known as an instantiated
graph \(G(x^i_W)=(V, E, x^i_W)\), namely, a graph \(G=(V,E)\) whose nodes
\(W\subseteq V\) are associated to the assignment \(x^i_W\in\val(W)\)
\cite{hojsgaard2000yggdrasil}. Unfortunately, a set of features is not easily
interpretable as a single graph, because we cannot efficiently verify
independence assertions, since we are required to check all the features in
\F.

A graph representation for overcoming the previous limitations is canonical
models \cite{edera2014a}. These models are a proper subset of the CSI models
\cite{hojsgaard2000yggdrasil,hojsgaard2004statistical}, which can capture
context-specific independences in a more interpretable way than a set of
features. A canonical model \(\bar\G\) is a pair \((\G, \X)\), where \(\G\) is
a collection of instantiated graphs of the form
\(\G=\{G(x^i)\in\G~:~x^i\in\X\subseteq\val(V)\}\), and \X\ is a set of
canonical assignments. These instantiated graphs are called \emph{canonical
  graphs}, because every graph \(G(x^i)\) is associated to a canonical
assignment \(x^i\in\val(V)\). In contrast to a single graph \(G\), a canonical
model requires several canonical graphs for capturing both conditional and
context-specific independences. For instance, let us suppose that we want to
encode the context-specific independence \ci{X_a}{X_b}{x_w} in a canonical
model \(\bar\G\). By Definition~\ref{sec:cont-spec-cond}, this independence
implies a set of independences of the form~\ci{x_a}{x_b}{x_w}, for all the
assignments \(x_a,x_b\in\val(a),\val(b)\). Then, each independence
\ci{x_a}{x_b}{x_w} is captured by a particular \(G(x^i)\in\G\), one whose
context \(x^i\) satisfies: \(x_a^i=x_a\), \(x_b^i=x_b\), and \(x_w^i=x_w\).

\subsection{Markov networks}
\label{sec:markov-networks}

A Markov network is a parametric model for representing probability
distributions in a compact way. This model is defined by a structure and a set
of potential functions \(\{\phi_k(X_{D_k})\}_k\), where \(\phi_k:
\val(D_k)\mapsto\mathbb{R}^+\), and \(X_{D_k}\subseteq X\) is known as the
\emph{scope} of \(\phi_k\). For discrete domains, a usual representation of
the potential functions is a table-based function. Markov networks can
represent a very important class of probability distributions called
\emph{Gibbs distributions}, whose functional form is as follows: \(
p(X=x)=\frac{1}{Z}\prod_k \phi_k(x_{D_k})\), where \(Z\) is a global constant,
called \emph{partition function}, that guarantees the normalization of the
product.
A Gibbs distribution \(p(X)\) \emph{factorizes} over a graph \(G\), if any
scope \(X_{D_k}\) corresponds to a complete subgraph \(D_k\) (a.k.a.
\emph{clique}) of the graph \(G\). Without loss of generality, the Gibbs
distribution is often factorized by using the \emph{maximum cliques} of the
graph \(G\). For positive distributions, one important theoretical result
states the converse \cite{Hammersley_Clifford_1971}, that is, \(p(X)\) can be
represented as a Gibbs distribution (Markov network) that factorizes over
\(G\), if \(G\) is an \emph{I-map}\footnote{A structure is an I-map for
  \(p(X)\) if every independence described by the structure holds in
  \(p(X)\).} for \(p(X)\). As a result, given a positive Gibbs distribution
\(p(X)\), it can be shown that every influence on any variable \(X_a\in X\)
can be blocked by conditioning on its Markov blanket \(\mbg{a}{G}\), formally:
\( p(X_a\vert X_{V\setminus\{a\}}) = p(X_a\vert X_{\mb{a}})\)\footnote{We
  further refer the readers to Section~3.2.1 in \cite{lauritzen1996graphical}
  and Section~4.3.2 in \cite{koller09} for more details about Markov
  properties on undirected graphs.}. Interestingly, an extension of the
previous property provides a criterion for determining the presence or absence
of any edge \((a,b)\) in an I-map graph \(G\) as follows
\cite[Theorem~1]{schluter2014ibmap}:

\begin{proposition}
\label{prop:criterion}
Let \(p(X)\) be a positive Gibbs distribution. Then, for any \(a\in V\):

  \begin{enumerate}
  \item the set of assertions
    \(\{\ci{X_a}{X_b}{X_{\mb{a}\setminus\{b\}}}~\colon~ b\in\mb{a}\}\) is
    \emph{false} in \(p(X)\), presence of an edge \((a,b)\), iff each
    assertion satisfies \(p(X_a,X_b\vert X_{\mb{a}}) \neq p(X_a\vert
    X_{\mb{a}})\cdot p(X_b\vert X_{\mb{a}})\).

  \item the set of assertions \(\{\ci{X_a}{X_b}{X_{\mb{a}}}~\colon~
    b\notin\mb{a}\}\) is \emph{true} in \(p(X)\), absence of an edge
    \((a,b)\), iff each assertion satisfies \(p(X_a,X_b\vert X_{\mb{a}}) =
    p(X_a\vert X_{\mb{a}})\cdot p(X_b\vert X_{\mb{a}})\).
  \end{enumerate}

\end{proposition}


Although a Gibbs distribution makes the structure explicit, it encodes the
potential functions as a table-based function, obscuring finer-grained
structures such as context-specific independences \cite{koller09}. For this
reason, a commonly used representation of a Markov network is the
\emph{log-linear model} defined as \( p(x) = \frac{1}{Z} \exp \left\{~
  \sum_k\sum_i w_{i,k} f^i_k(x_{D_k}) \right\}.\) A log-linear model can be
constructed from a Gibbs distribution as follows: for the \(i\)th row of the
table-based potential function \(\phi_k\), an indicator function
\(f^i_k(\cdot)\) is defined whose weight is \(w_{i,k}=\log\phi_k(x^i_{D_k})\).

\section{Context-Specific Grow-Shrink algorithm}
\label{sec:csgs}

In this section we present CSGS (Context-Specific Grow-Shrink), a knowledge
discovery algorithm for learning the structure of Markov networks by using
canonical models as structure representation. The design of CSGS was inspired
by the search strategy used by CSPC for learning canonical models
\cite{edera2014a}, and the GS search strategy for learning graphs
\cite{margaritis00,bromberg2009efficient}. Therefore, CSGS obtains a canonical
model by learning a collection \G\ of mutually independent canonical graphs,
where each canonical graph \(G(x^i)\in\G\) is learned by using the GS
strategy. More precisely, GS obtains a graph in two steps: first, it
\emph{generalizes} an initial very specific graph (one that makes many
independence assumptions) by adding edges. Then, the resulting graph is
\emph{specialized} by removing spurious edges. In sum,
Algorithm~\ref{alg:csgs:overview} shows an overview of CSGS. In
line~\ref{alg:csgs:setofcanonical}~and~\ref{alg:csgs:canonicalmodel}, CSGS
defines an initial specific canonical model from a set of canonical
assignments \X. Subsequently,
lines~\ref{alg:csgs:contexts}~and~\ref{alg:csgs:gs} construct each canonical
graph \(G(x^i)\in\G\) by using the GS strategy. For determining the presence
or absence of an edge, CSGS uses Proposition~\ref{prop:criterion} as
criterion. The validation of this criterion is realized by eliciting
context-specific independences from data in a similar way to CSPC
\cite[Section 4.3]{edera2014a}. Finally, in a similar fashion to CSPC
\cite[Section 4.4]{edera2014a}, CSGS uses the resulting canonical model
\(\bar\G\) for generating a set \F\ of features in order to enable us to use
standard software packages for performing weight learning and inference. The
remaining of this section is structured by using the key elements of CSGS:
\textit{i)} Section~\ref{sec:init-canon-model} describes how the initial
canonical model is defined; \textit{ii)}
Section~\ref{sec:grow-shrink-strategy} presents the GS strategy for obtaining
the canonical graphs; and \textit{iii)} Section~\ref{sec:asympt-compl}
concludes analyzing the time complexity of CSGS.

\vspace*{-.5cm}
\input{alg-overview-csgs.tex}
\vspace*{-1.2cm}

\subsection{Initial canonical model}
\label{sec:init-canon-model}

The definition of the initial canonical model consists, firstly, in the set of
canonical assignments \X. In a similar fashion to CSPC \cite{edera2014a}, this
set is composed by the unique training examples in \D. This definition is the
consequence of using the \emph{data-driven approach}, that is, we use only
contexts that appear in data, and for the remaining contexts which do not
appear in the data, we assume that they are improbable due to the lack of
other information. Lastly, once \X\ is defined, we associate the most specific
graph \(G(x^i)\) to each context \(x^i\in\X\), namely, the empty graph. As a
result, in each initial canonical graph, every Markov blanket is empty. The
idea behind the GS strategy is to add edges, thus adding nodes to each
blanket.

\subsection{Grow-Shrink strategy for learning canonical graphs}
\label{sec:grow-shrink-strategy}

CSGS uses the GS strategy under the \emph{local-to-global approach}
\cite{schluter2014ibmap,JMLR:v15:lowd14a}. In this approach, the structure is
obtained by constructing each Markov blanket \(\mb{a}, a\in V\) in turn. In
this manner, for each node \(a\), the strategy GS determines the Markov
blanket \mb{a} in two phases: the \emph{grow phase} and the \emph{shrink
  phase}. The grow phase adds a new edge \((a,b)\) to \(E\) as long as
Proposition~\ref{prop:criterion}.1 is satisfied in data. However, due to the
node ordering used \cite{margaritis00,bromberg2009efficient}, the grow phase
can add nodes that are outside of the blanket, resulting in spurious
edges. For this reason, the shrink phase removes an edge \((a,b)\in E\) as
long as Proposition~\ref{prop:criterion}.2 is satisfied in
data. Algorithm~\ref{alg:csgs:inner} shows a more detailed description of the
construction of the canonical graph \(G(x^i)\). Initially, the canonical graph
\(G(x^i)\) is empty, then it is generalized by using the local-to-global
approach shown in the loop of line~\ref{alg:csgs:main-loop}. In this loop, the
two steps of GS are performed: the grow phase, starting in
line~\ref{alg:csgs:grow}; and the shrink phase, starting in
line~\ref{alg:csgs:shrink}. In each iteration of the main loop,
line~\ref{alg:csgs:add}~and~\ref{alg:csgs:remove} change the Markov blanket by
adding/removing new edges to the current set \(E\) of edges. Once the main
loop has finished, the Markov blankets of each node are obtained and, in
consequence, the resulting canonical graph encodes context-specific
independences.

\input{alg-inner-csgs.tex}
\vspace*{-1.4cm}

\subsection{Asymptotic complexity}
\label{sec:asympt-compl}

As is usual in knowledge discovery algorithms, we analyze the complexity of
CSGS by determining the number of independence tests performed for
constructing a structure from data. Let \(m\) be the number of unique examples
in the dataset, the complexity of performing a test is linear in
\(m\). However, this cost can be particularly high if \(m\) is large. In our
implementation of CSGS, we reduce this cost by using
ADTree~\cite{moore1998cached}. We assume that nodes in
line~\ref{alg:csgs:main-loop} in Algorithm~\ref{alg:csgs:inner} are taken in
an unspecified but fixed order, and we bound the maximum degree of a node to
\(k=\argmax_{G(x^i)\in\G}\argmax_{a\in V}(|\mbg{a}{G(x^i)}|)\). Let \(n\) be
the number of variables, and let \(G(x^i)=(V,E,x^i)\) be an empty canonical graph,
we can decompose the analysis into the number of tests performed by grow and
shrink phases. In the grow phase, a test is performed for each edge
\((a,b)\notin E\), resulting in \BigO{n^2} tests. At the end of the grow
phase, the size of a blanket is \(k\) at worst, thus shrink phase performs
\BigO{nk} tests. Additionally, Algorithm~\ref{alg:csgs:overview} performs the
GS strategy \(m\) times, one per each initial canonical graph. Therefore, the
total complexity is \BigO{m(n^2 + nk)} independence tests.







\section{Empirical evaluation}
\label{sec:empirical-evaluation}

This section shows experimental results obtained from the structures learned
by CSGS and several structure learning algorithms on synthetic
datasets. Basically, the goals of our experiments remark the greatly practical
utility of our algorithm in a two-fold manner. First, we compare the accuracy of the
structures learned by CSGS and CSPC, as well as by other state-of-the-art
structure learners. Second, we compare the computational complexity between
CSGS and CSPC. For evaluating the accuracy of the learned structures, we use
the underlying distributions that were sampled to generate the synthetic
datasets; since there is a direct correlation between the correctness of the
structure and the accuracy of the distribution\cite{Hammersley_Clifford_1971},
the accuracy of a structure can be measured by 
comparing the similarity between the learned and underlying distributions. On
the other hand, for evaluating the computational complexity, we report the
number of tests performed for constructing the structures\footnote{Additional
  empirical results are available in the online appendix
  \url{http://dharma.frm.utn.edu.ar/papers/iberamia14/supplementary-information-on-csgs.pdf}}. Lastly,
an open source implementation of CSGS algorithm as well as the synthetic datasets
used in this section are publicly available\footnote{\url{http://dharma.frm.utn.edu.ar/papers/iberamia14}}.

\subsection{Datasets} \label{sec:datasets}

The datasets of our experiment are used in \cite{edera2013a,edera2014a} and
were sampled from Markov networks with context-specific independences for
different \(n\) numbers of variables that range from 6 to 9, varying their
sizes from 20 to 100k datapoints. For each \(n\), 10 datasets were sampled
from 10 different Markov networks with fixed structure but randomly choosing
their weights. For more details, we refer the readers
to~\cite[Appendix~B]{edera2014a}. Roughly speaking, the underlying structure
of these models encodes independence assertions of the form
\ci{X_a}{X_b}{x^1_w} for all pairs \(a,b\in V\setminus\{w\}\), becoming
dependent when \(X_w=x^0_w\). In this way, the underlying structure can be
seen as two instantiated graphs: a fully connected graph \(G(x^0_w)\), and a
star graph \(G(x^1_w)\) whose central node is \(x^1_w\). Despite the
simplicity of this structure, this cannot be correctly captured by using a
single graph, yet it can be captured by sets of features or canonical
models. On the other hand, as the maximum degree of the underlying structure
is equal to \(n\), learning the structure is a challenging problem
\cite{edera2013a,schluter2014ibmap}. The generated datasets are partitioned
into: a \emph{training set} (70\%) and a \emph{validation set}
(30\%). The validation set is used by density estimation algorithms to set their
tuning parameters. Specifically, they use different tuning parameters for
learning several structures from the training set, selecting one whose
pseudo-likelihood on the validation set is maximum. In contrast, CSGS, CSPC,
GSMN and IBMAP-HC algorithms do not use tuning parameters, thus they learn
structures by using the whole dataset, i.e. the union of training and
validation sets.

\subsection{Methodology}

In this subsection we explain the methodology used for evaluating our approach
against several structure learning algorithms. First, we explain which
structure learning algorithms are used as competitors and their configuration
settings, and then we describe the method used for measuring the accuracies of
the learned structures: \emph{Kullback-Leibler divergence} (KL)
\cite[Appendix~A]{koller09}.

CSGS is compared against CSPC (Context-Specific Parent and Children) algorithm
and two representative algorithms for knowledge discovery and density
estimation goals. 
The knowledge discovery algorithms are: GSMN (Grow-Shrink Markov Network
learning algorithm) \cite{bromberg2009efficient}, and IBMAP-HC (IBMAP
Hill-Climbing)
\cite{schluter2014ibmap}. 
For a fair comparison, we use the Pearson's $\chi^2$ as the statistical
independent test with a significance level of \(0.05\) for CSGS, CSPC and
GSMN, but not for IBMAP-HC which only works with the Bayesian statistical test
with a threshold equal to \(0.5\). On the other hand, the density estimation
algorithms are: GSSL (Generate Select Structure Learning)
\cite{van2012markov}, and DTSL (Decision Tree Structure Learner)
\cite{JMLR:v15:lowd14a}. For a fair comparison, we replicate the recommended
tuning parameters for both algorithms detailed in \cite{van2012markov}, and
\cite{lowd2010learning}, respectively. 
%
%
KL divergence is a ``distance measure'' widely used to evaluate how similar
are two distributions. Thus, using the learned structures, we obtain Markov
networks by learning their weights with pseudo-likelihood\footnote{Weight
  learning was performed using version 0.5.0 of the Libra toolkit
  (\url{http://libra.cs.uoregon.edu/}).}, measuring their KL divergences with respect
to the underlying distribution. Lower values of KL divergence indicate better
accuracy.

\subsection{Results of experimentation}

\begin{figure*}[ht!]
  \centering
  \includegraphics[width=0.495\textwidth]{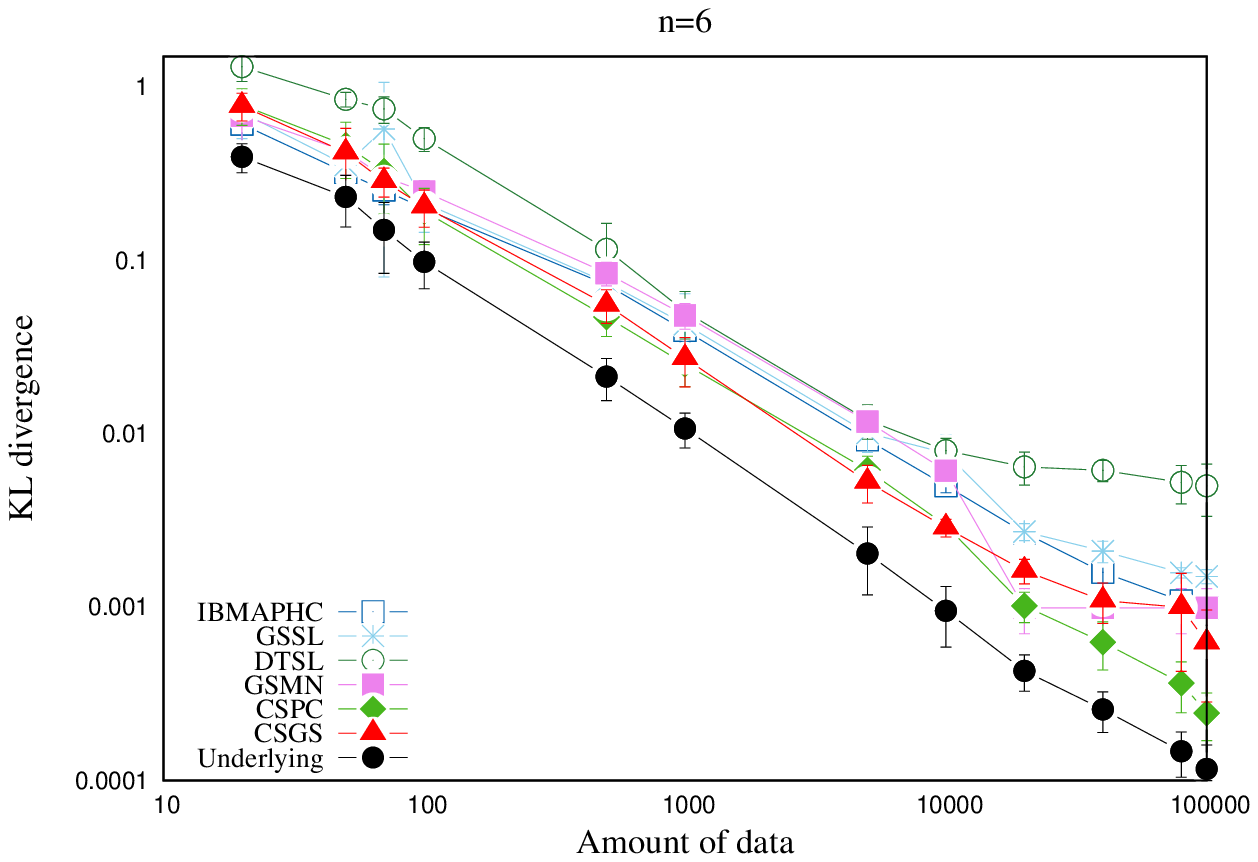}
  \includegraphics[width=0.495\textwidth]{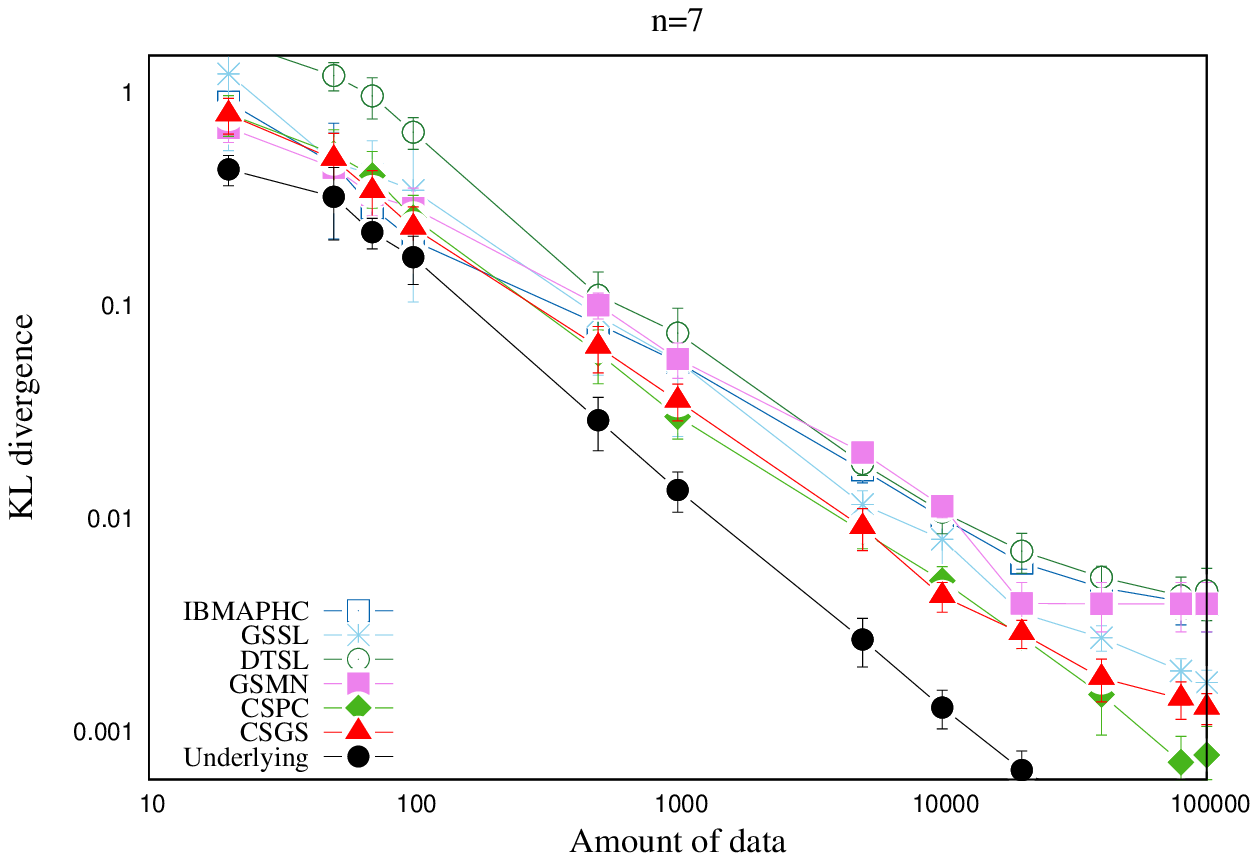}

  \includegraphics[width=0.495\textwidth]{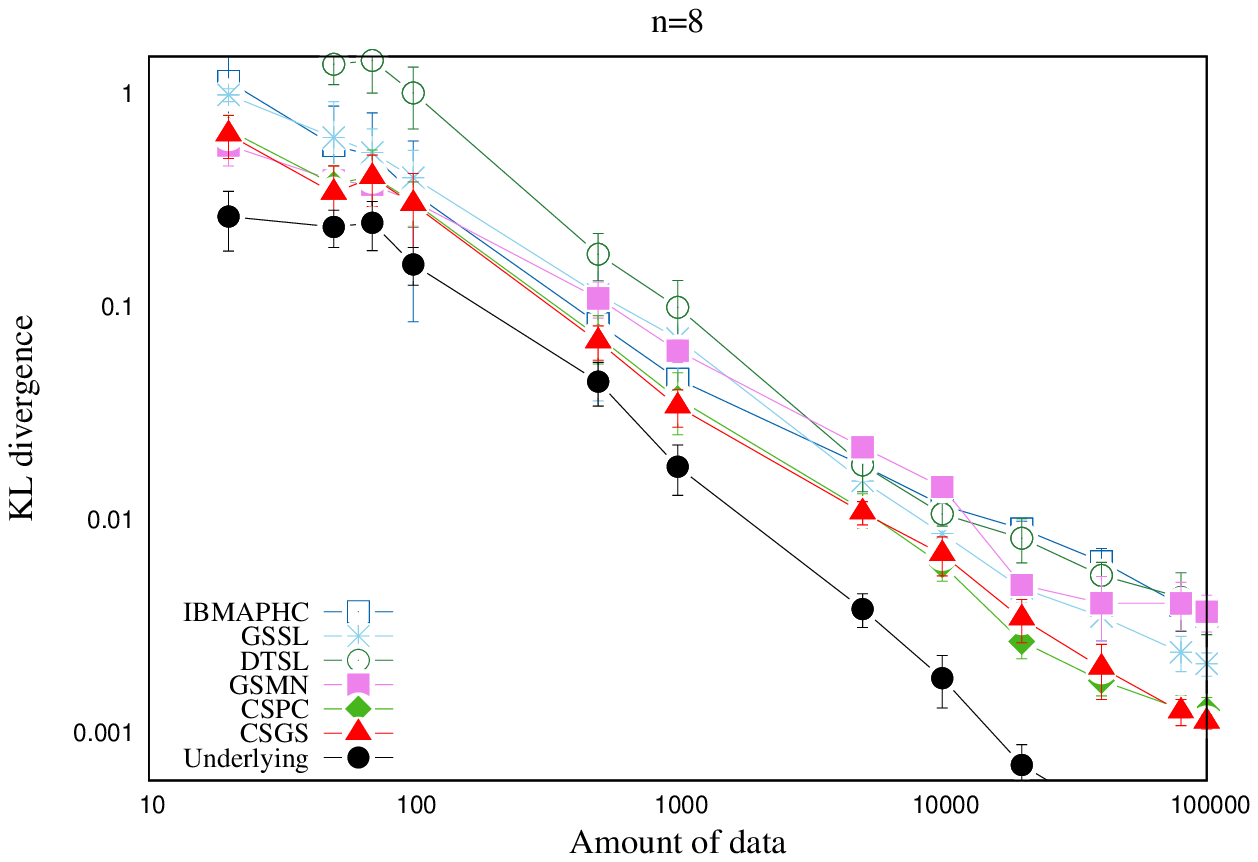}
  \includegraphics[width=0.495\textwidth]{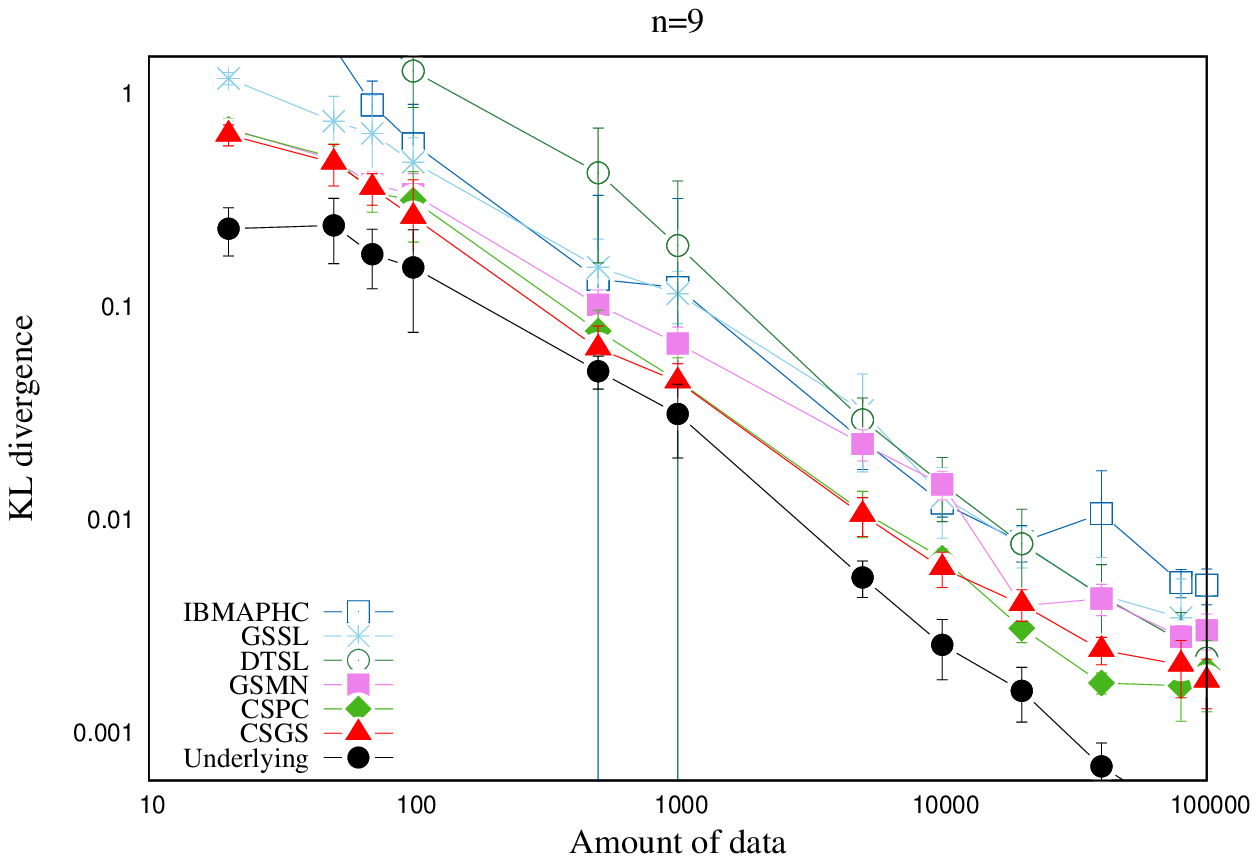}

  \caption{KL divergences over increasing amounts of data for structures
    learned by several learning algorithms. For comparison reasons, the KL
    divergence of the underlying structure is shown. Every point represents
    the average and standard deviation over ten datasets with a fixed
    size. Lower values indicate better accuracy.}
  \label{fig:kl}
\end{figure*}

\begin{figure*}[ht!]
  \centering
  \includegraphics[width=0.48\textwidth]{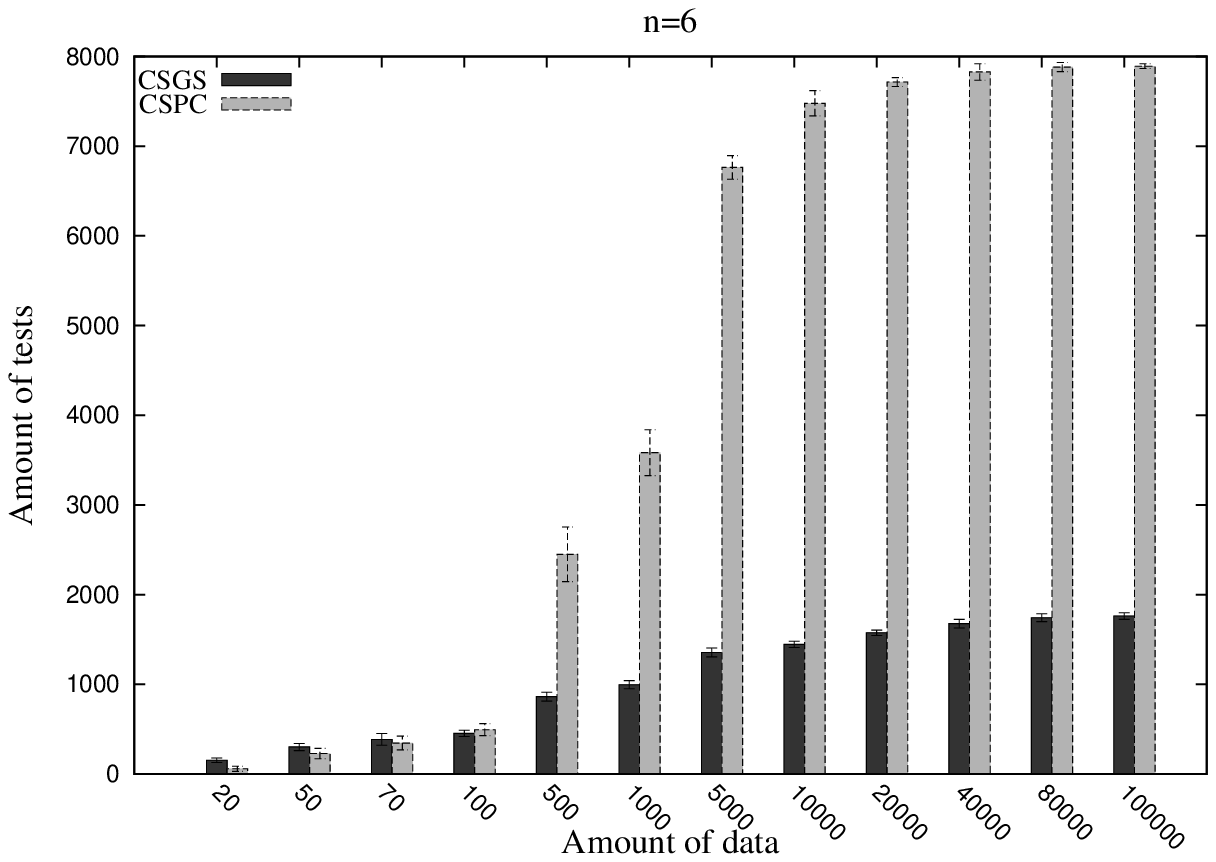}
  \includegraphics[width=0.48\textwidth]{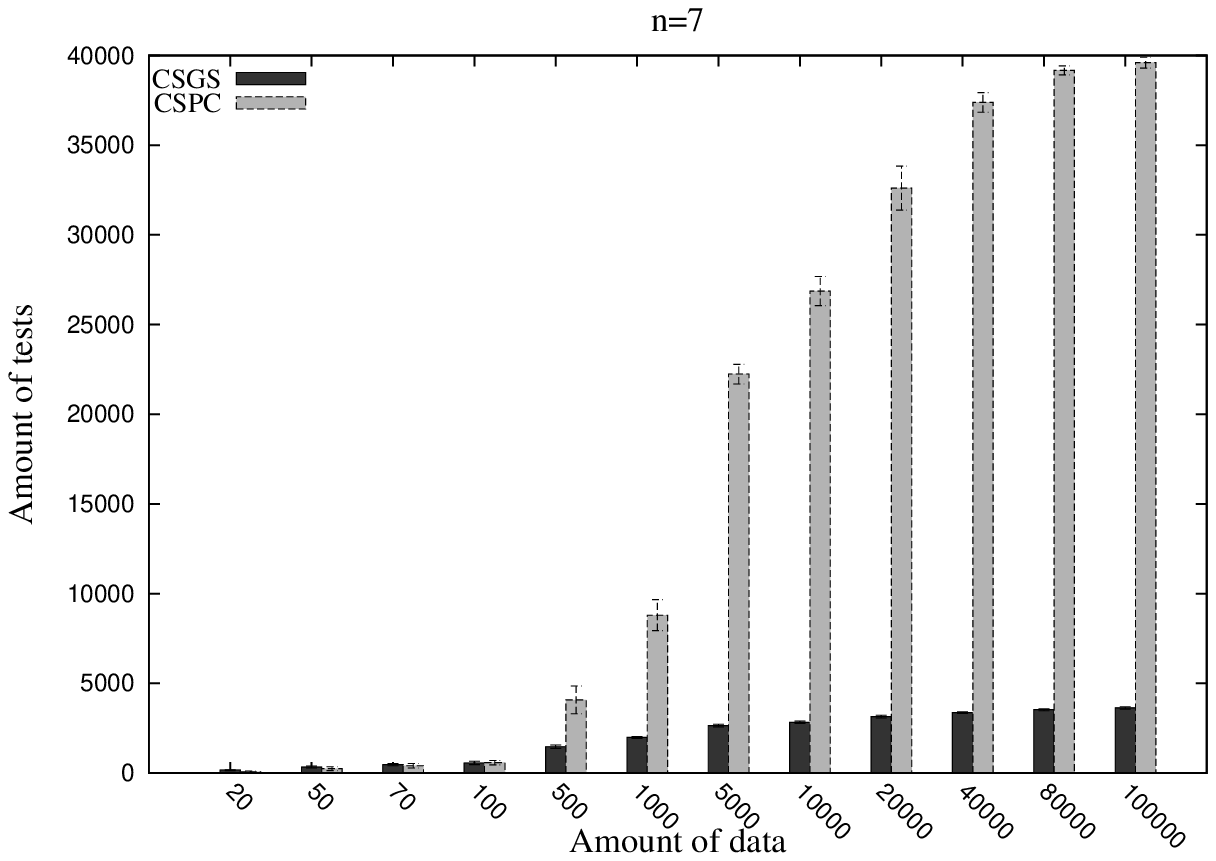}

  \includegraphics[width=0.48\textwidth]{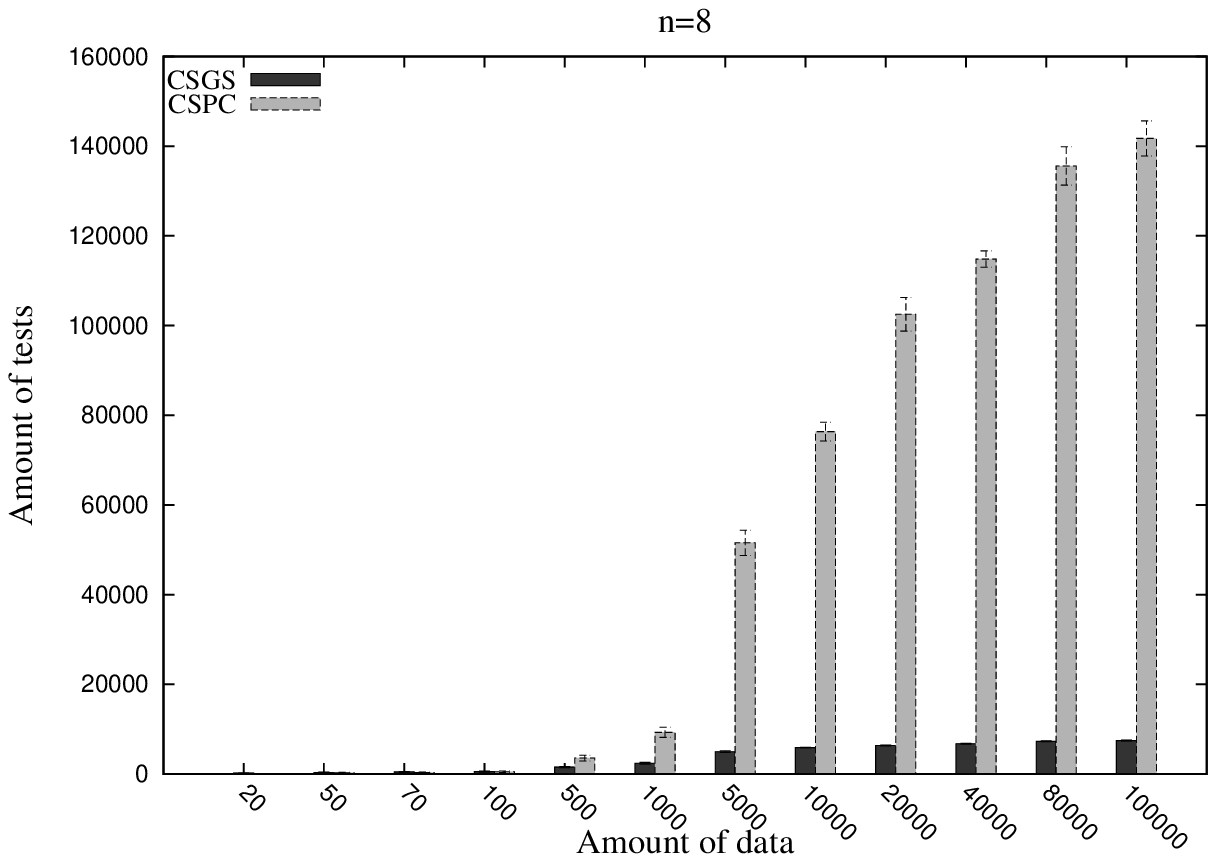}
  \includegraphics[width=0.48\textwidth]{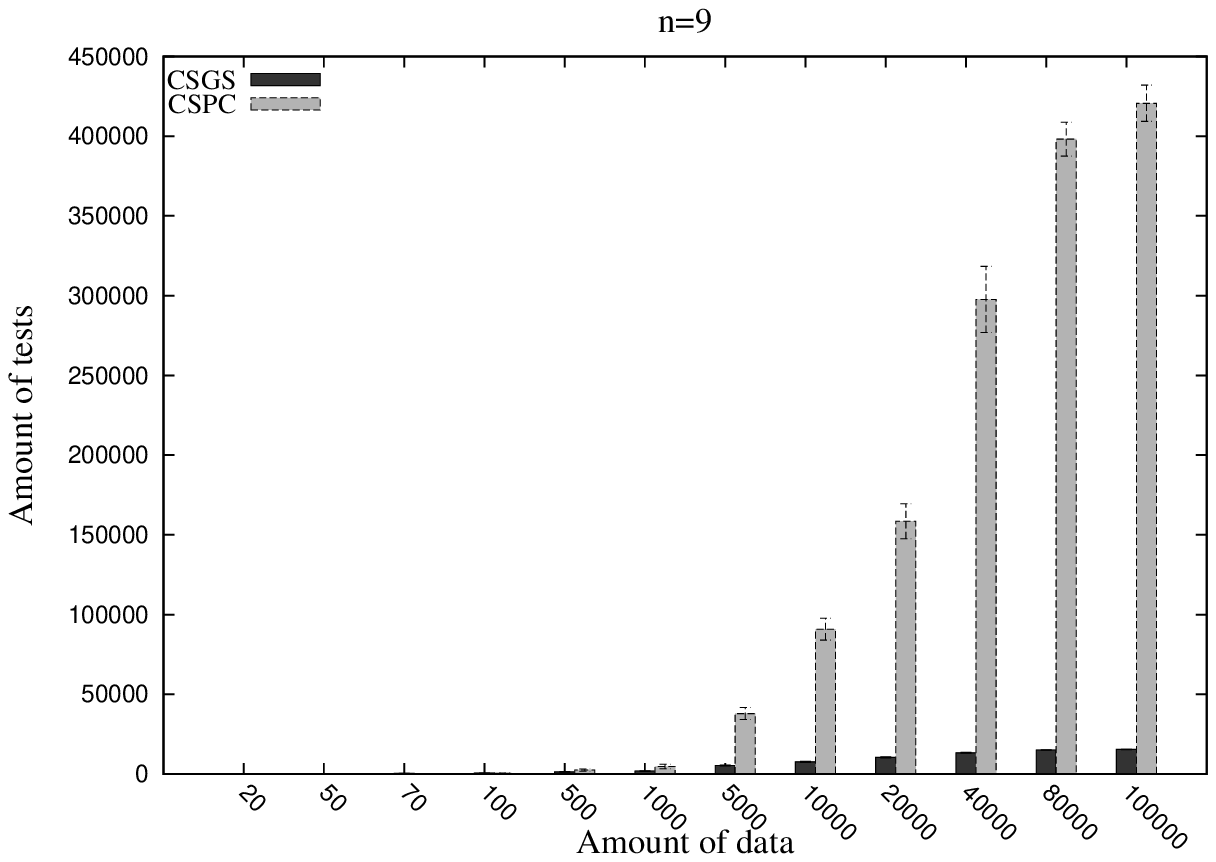}

  \caption{Number of tests performed by CSGS and CSPC for learning structures
    over increasing amounts of data. Every bar represents the average and
    standard deviation over ten datasets with a fixed size.}
  \label{fig:nt}
\end{figure*}

Figure~\ref{fig:kl} presents the KL divergences computed from the structures
learned by the different algorithms. For comparison reasons,
Figure~\ref{fig:kl} also shows the KL divergence computed by using a Markov
network whose structure is the underlying one, showing the best KL divergence
that can be obtained. In these results, we can see three important
trends. First, the structures learned by CSGS reach similar divergences in
most cases to CSPC. Second, in most cases, the divergences obtained by CSGS
and CSPC are better than those obtained by the other structure
learners. Finally, the divergences of CSGS and CSPC are closer to the
divergences obtained by the underlying structure. These trends allow us to
conclude that the structures learned by CSGS and CSPC can encode the
context-specific independences present in data, resulting in Markov networks
more accurate than those obtained by the remaining
algorithms. Figure~\ref{fig:nt} presents the number of tests performed by CSGS
and CSPC for learning the structures used previously for computing the KL
divergences. As shown, the number of tests performed by CSGS is smaller than
those performed by CSPC. The difference between both dramatically increases
as data increases. These results show the great impact of using the GS
strategy for learning canonical models. In conclusion, the results shown in
both figures show that CSGS is an efficient alternative to CSPC for learning
canonical models.

\section{Conclusions and future work}
\label{sec:concl-future-work}

In this work we presented CSGS, a new knowledge discovery algorithm for
learning Markov network structures by using canonical models. CSGS is similar
to the CSPC algorithm \cite{edera2014a}, except that CSGS uses an alternative
search strategy called Grow-Shrink \cite{margaritis00,bromberg2009efficient},
that avoids performing unnecessary independence tests. We evaluated our
algorithm against CSPC and several state-of-the-art learning algorithms on
synthetic datasets. In our results, CSGS learned structures with similar
accuracy to CSPC but performing a reduced number of tests. The directions of
future work are focused on further reducing the computational complexity and
improving the quality of the learned structures using alternative search
strategies. For instance, IBMAP-HC on the side of knowledge discovery
algorithms \cite{schluter2014ibmap}, and GSSL on the side of density
estimation algorithms \cite{van2012markov}.


\end{document}

%% file: alg-overview-csgs.tex
\begin{algorithm}[!h]\small
  \scriptsize
  \KwIn{domain \(V\), dataset \D }

  \BlankLine
  \(\X\leftarrow\) Define the set of canonical assignments  \nllabel{alg:csgs:setofcanonical}

  \(\G\leftarrow\) Define a set of initial graphs \(\{G(x^i) \colon
  x^i\in\X\}\) \nllabel{alg:csgs:canonicalmodel}

  \ForEach{\(G(x^i)\in\G\)}
  { \nllabel{alg:csgs:contexts}
    \(G(x^i) \leftarrow\) GS(\(G(x^i)\), \D) \nllabel{alg:csgs:gs}
  }

  \(\F \leftarrow\) Feature generation from \(\bar\G=(\G,\X)\) \nllabel{alg:csgs:features}


  \caption{\textsc{Overview of CSGS}}
  \label{alg:csgs:overview}
\end{algorithm}

%% file: alg-inner-csgs.tex
\begin{algorithm}[h]\small
  \scriptsize
  \KwIn{graph \(G(x^i)=(V,E,x^i)\), dataset \D}

  \BlankLine

    \ForEach{node \(a\in V\)} { \nllabel{alg:csgs:main-loop}



      \ForEach {node $b \in V\setminus ( {\mbg{a}{G(x^i)}} \cup \{a\})$ } { \nllabel{alg:csgs:grow}

          \If{$\ci{X_a}{X_b}{x^i_{\mbg{a}{G(x^i)}}}$ is false in \D} { \nllabel{alg:csgs:test}

             \(E\leftarrow E\cup(a,b)\) \nllabel{alg:csgs:add}
           }
          }

    \ForEach {$b \in  \mb{a}$  } { \nllabel{alg:csgs:shrink}

      \If{$\ci{X_a}{X_b}{x^i_{\mbg{a}{G(x^i)}\setminus \{b\}}}$ is true in \D} { \nllabel{alg:csgs:test2}

            \(E\leftarrow E\setminus(a,b)\) \nllabel{alg:csgs:remove}

            }

            }

    }

  \Return{\(G(x^i)\)}  \nllabel{alg:csgs:return2}

  \caption{\textsc{GS strategy}}
  \label{alg:csgs:inner}
\end{algorithm}